\documentclass[10pt, conference, compsocconf]{IEEEtran}
\usepackage{xcolor}
\usepackage[square,numbers]{natbib}
\usepackage{xspace}
\usepackage{hyperref}       
\usepackage{caption}
\usepackage{subcaption}
\usepackage{graphicx}

\ifCLASSINFOpdf
\else
\fi
\usepackage[cmex10]{amsmath}
\usepackage{url}

\DeclareMathOperator*{\argmin}{arg\,min}
\newcommand{\eg}{e.g.\xspace}
\newcommand{\ie}{i.e.\xspace}

\newcommand{\niid}{non-i.i.d.\xspace}

\DeclareMathOperator{\sign}{sgn}

\begin{document}
%
\title{The Impact of Data Distribution on Fairness and Robustness \\ in Federated Learning}


\author{\IEEEauthorblockN{Mustafa Safa Ozdayi}
\IEEEauthorblockA{Department of Computer Science \\
The University of Texas at Dallas \\
Richardson, TX, USA \\
mustafa.ozdayi@utdallas.edu}
\and
\IEEEauthorblockN{Murat Kantarcioglu}
\IEEEauthorblockA{Department of Computer Science \\
The University of Texas at Dallas\\
Richardson, TX, USA\\
muratk@utdallas.edu}
}


%


\maketitle

\begin{abstract}
Federated Learning (FL) is a distributed machine learning protocol that allows a set of agents to collaboratively train a model without sharing their datasets. This makes FL particularly suitable for settings where data privacy is desired. However, it has been observed that the performance of FL is closely related to the similarity of the local data distributions of agents. Particularly, as the data distributions of agents differ, the accuracy of the trained models drop. In this work, we look at how variations in local data distributions affect the \emph{fairness} and the \emph{robustness} properties of the trained models in addition to the accuracy. Our experimental results indicate that, the trained models exhibit higher bias, and become more susceptible to attacks as local data distributions differ.  Importantly, the degradation in the fairness, and robustness can be much more severe than the accuracy. Therefore, we reveal that small variations that have little impact on the accuracy could still be important if the trained model is to be deployed in a fairness/security critical context.
\end{abstract}

%
\IEEEpeerreviewmaketitle

\section{Introduction}
Federated Learning (FL)~\cite{fed-learning:google} is a distributed machine learning protocol. Through FL, a set of agents can collaboratively train a model without sharing their data with each other, or any other third party. This makes FL suitable to settings where data privacy is desired. In this regard, FL differs from the traditional distributed learning setting in which data is first centralized at a place, and then distributed to the agents~\cite{dean2012large,li2014scaling}. 

The decentralized nature of datasets in FL brings some unique challenges. In particular, several works have observed that the accuracy of the trained models can drop significantly when the data distributions of the agents are different (the so-called \niid setting), and have tried to mitigate this negative effect~\cite{zhao2018federated,fedNonIID2, li2020federated, ozdayi2020improving, shoham2019overcoming, hsieh2020non}. 

To to best of our knowledge, the existing literature have almost exclusively analyzed the impact of data distributions on the accuracy. In contrast, in this work, we look at how data distributions of agents affect the \emph{fairness} and \emph{robustness} of the trained models.
That is, we ask, \emph{how does the fairness and robustness properties of the trained models change with the distribution of agents' datasets?}
In this work, we quantify the bias exhibited by the model based on the class-wise accuracies,
(see Section~\ref{sec:expFairness}), and quantify the robustness based on the resilience against \emph{model poisoning} attacks (see Section~\ref{sec:bd_attacks} and~\ref{sec:expRobustness}).

Briefly, our analysis reveals that, just as for accuracy, fairness and robustness of the trained models degrade as the distribution of local datasets differ. More importantly though, we show that this degradation can be much more severe than the degradation of the accuracy. We believe these observations can be important in several contexts. For example, the existing defenses in the FL literature are usually tested against different ratios of adversarial agents. However, as we show that the models become more susceptible to attacks as local data distributions change, it is also important to test the effectiveness of the defenses under different local data distributions.

The rest of the paper is organized as follows: in Section~\ref{sec:background}, we provide the necessary background and discuss the related works. In Section~\ref{sec:dataDist}, we provide our experimental results, and show how robustness and fairness of the trained models change with the local data distributions. Finally in Section~\ref{sec:conc}, we discuss our results and provide some concluding remarks.

\section{Background and Related Work}\label{sec:background}
In this section, we provide the necessary background on FL and explain model poisoning attacks.
\subsection{Federated Learning (FL)}
At a high level, FL  is multi-round protocol between an aggregation server and a set of agents in which agents jointly train a model. Formally, participating agents try to minimize the average of their loss functions,
$$
\argmin_{w \in R^d} f(w) = \frac{1}{K}\sum_{k=1}^K f_k(w),
$$
where $f_k$ is the loss function of $k$th agent. For example, for neural networks, $f_k$ is typically empirical risk minimization under a loss function $L$ such as cross-entropy, \ie,

$$
f_k(w) = \frac{1}{n_k} \sum_{j=1}^{n_k} L(x_j, y_j; w),
$$
with $n_k$ being the total number of samples in agent's dataset and $(x_j,y_j)$ being the $j$th sample.

Concretely, FL protocol is executed as follows: at round $t$, server samples a subset of agents $S_t$, and sends them $w_t$, the model weights for the current round. Upon receiving $w_t$,  $k$th agent initializes his model with the received weight, and trains for some number of iterations, \eg, via stochastic gradient descent (SGD), and ends up with weights $w_t^k$. The agent then computes his update as $\Delta_t^k = w_t^k - w_t$, and sends it back to the server. Upon receiving the update of every agent in $S_t$, server computes the weights for the next round by aggregating the updates with an aggregation function $\mathbf{g} \colon R^{|S_t| \times d } \rightarrow R^d$ and adding the result to $w_t$. That is,
$w_{t+1} = w_t + \eta\cdot g( \{ \Delta_t \} )$ where $\{ \Delta_t \} = \cup_{k \in S_t}\Delta_t^k$, and $\eta$ is the server's learning rate.
For example, original FL paper~\cite{fed-learning:google} and many subsequent papers on FL~\cite{arxiv:2018:fedlens,arxiv:2018:backdoor,sun2019really,bonawitz2016practical,geyer2017differentially} consider weighted averaging to aggregate updates. In this context, this aggregation is referred as Federated Averaging (FedAvg), and yields the following update rule,
\begin{equation*}
w_{t+1} = w_t + \eta \frac{\sum_{k \in S_t} n_k \cdot \Delta_t^k}{\sum_{k \in S_t} n_k}.
\label{eqn:fedavg}
\end{equation*}
In practice, rounds can go on indefinitely, as new agents can keep joining the protocol, or until the model reaches some desired performance  metric (\eg, accuracy) on a validation dataset maintained by the server.

It has been shown that models trained via FL can perform better than locally trained models at agents' side in various settings ~\cite{fed-learning:google, keyboard}. 
In contrast, as noted before, it has also been observed that the performance of FL drops drastically when local data distributions of agents differ significantly, \ie, when data is distributed in a \niid fashion among agents~\cite{zhao2018federated,fedNonIID2, li2020federated, ozdayi2020improving, shoham2019overcoming, hsieh2020non}.

\subsection{Backdoor Attacks and Model Poisoning}
\label{sec:bd_attacks}
Training time attacks against machine learning models can roughly be classified into two categories: targeted~\cite{arxiv:2018:fedlens,arxiv:2018:backdoor, chen2017targeted,liu2017trojaning}, and untargeted attacks~\cite{blanchard2017machine,bernstein2018signsgd}.
In untargeted attacks, the adversarial task is to make the model converge to a sub-optimal minima or to make the model completely diverge. Such attacks have been also referred as \emph{convergence attacks}, and to some extend, they are easily detectable by observing the model's accuracy on a validation data. 

On the other hand, in targeted attacks, adversary wants the model to misclassify only a set of chosen samples with minimally affecting its performance on the main task. Such targeted attacks are also known as \emph{backdoor attacks}. A prominent way of carrying backdoor attacks is through \emph{trojans}~\cite{chen2017targeted,liu2017trojaning}. A trojan is a carefully crafted pattern that is leveraged to cause the desired misclassification. For example, consider a classification task over cars and planes and let the adversarial task be making the model classify blue cars as plane. Then, adversary could craft a brand logo, put it on \emph{some} of the blue car samples in the training dataset, and only mislabel those as plane. Due to this attack, the attacked model would potentially learn to classify a blue car with the brand logo as a plane. At the inference time, adversary can present a blue car sample with the logo to the model to activate the backdoor. Ideally, since the model would behave correctly on blue cars that do not have the trojan, it would not be possible to detect the backdoor on a clean validation dataset.

In FL, the training data is decentralized and the aggregation server is only exposed to model updates. Given that, backdoor attacks are typically carried by constructing malicious updates. That is, adversary tries to create an update that encodes the backdoor in a way such that, when it is aggregated with other updates, the aggregated model exhibits the backdoor. This has been referred as \emph{model poisoning} attack  \cite{arxiv:2018:fedlens,arxiv:2018:backdoor,sun2019really}. For example, an adversary could control some of the participating agents in a FL instance and train their local models on trojaned datasets to construct malicious updates.
\section{Experiments}\label{sec:dataDist}
We now show how fairness and robustness of the trained models are impacted due to differences in local data distributions via experiments.
The general setting of our experiments are as follows: we are training models via FL for classification tasks among $10$ agents where in each around, an agent trains his local model for $2$ epochs with a batch size of $256$ with the Adam optimizer~\cite{kingma2014adam}. 
We train until the model converges on the training dataset, and do our measurements on a held-out test dataset.

We use two datasets: Fashion MNIST~\cite{xiao2017fashionmnist}, and CIFAR10~\cite{cifar10}. On Fashion MNIST, we use the LeNet model~\cite{lecun:1998:mnist}, and on CIFAR10, we use a 6-layer convolutional neural network, consisting of 3 convolutional and 3 fully-connected layers.

To simulate different data distributions, we distribute the initial datasets to agents using Dirichlet distribution with different concentration parameters 
(denoted as $\alpha$) as in~\cite{desai2021blockfla}. 
Higher values of $\alpha$ indicate the distribution of agents' local datasets over the classes in the dataset are more similar.

Our implementation is in PyTorch~\cite{PyTorch}, and our code is publicly available at \url{https://github.com/TinfoilHat0/Fairness-Robustness-in-FedLearning}. All the reported results are averaged over 5 runs, and are presented in mean $\pm$ std format.

\subsection{Fairness}\label{sec:expFairness}
We quantify the bias of the model as the difference between the highest and the lowest accuracy across the classes. That is, if $M_c$ is model $M's$ accuracy on class $c$, we have that,
$$
\text{Bias}_M = \text{max}_c \left(M_c \right ) - \text{min}_c \left( M_c \right),
$$
where the\emph{higher values for bias indicates the model is less fair across classes.}

Our results are presented in the Table~\ref{tab:fairness}. As can be seen from these results,  bias exhibited by the model increases as local dataset distributions differ more. Interestingly though, the rate of increase in the bias seems much higher than the degradation in the accuracy. For example, for Fashion MNIST dataset, we see that, as we go from $\alpha=1$ to $\alpha=0.25$, the overall accuracy drops by merely about $2\%$, but the bias of the model almost doubles.

\begin{table}[t]
\large
\centering
\caption{Effect of data distribution on fairness for Fashion MNIST (top), and CIFAR10 (bottom) datasets.}
\begin{tabular}{ c c c } 
 \hline
 Dirichlet $\alpha$ & Accuracy & Bias \\
 \hline
 $1.0$ & $90.08 \pm 0.17$ & $24.48 \pm 1.84$ \\
  $0.5$ & $89.06 \pm 0.31$ & $37.3 \pm 5.83$ \\
  $0.25$ & $88.02 \pm 1.81$ & $45.78 \pm 25.62$ \\
\end{tabular}
\begin{tabular}{ c c c } 
 \hline
 Dirichlet $\alpha$ & Accuracy & Bias \\
 \hline
 $1.0$ & $76.52 \pm 0.76$ & $29.03 \pm 2.3$ \\
 $0.5$ & $75.17 \pm 0.62$ & $35.6 \pm 6.04$ \\
 $0.25$ & $72.27 \pm 1.81$ & $42.6 \pm 14.79$ \\
\end{tabular}
\label{tab:fairness}
\end{table}

\subsection{Robustness}\label{sec:expRobustness}
\begin{figure}[t]
  \begin{subfigure}[b]{0.49\textwidth}
  \center
    \includegraphics[scale=0.4]{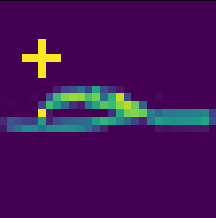}
    \includegraphics[scale=0.4]{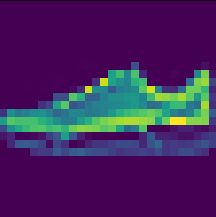}
     \caption{}
  \end{subfigure}
  \begin{subfigure}[b]{0.49\textwidth}
  \center
     \includegraphics[scale=0.4]{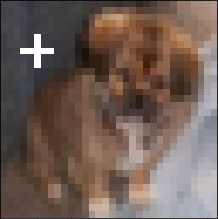}
     \includegraphics[scale=0.4]{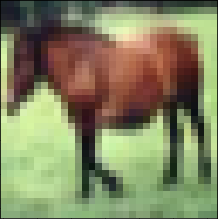}
     \caption{}
  \end{subfigure} 
  \caption{Samples from trojaned base classes and corresponding target classes. Trojan pattern is a 5-by-5 plus sign that is put to the top-left of objects. For Fashion MNIST (a), the backdoor task is to make model classify trojaned inputs as sneakers. For CIFAR10 (b), it is to make model classify trojaned samples as horses. For Fashion MNIST, we note that original images are in grayscale, and these figures are normalized as they appear in training/test datasets.}\label{fig:trojaned_samples}
\end{figure}

We now test how robustness of the model changes against \emph{model poisoning} attacks for different data distributions. In this experiment, we designate one of the agents (out of ten) as the adversary. We fix his dataset to 1000 samples uniformly sampled from the initial training dataset. The corrupt agent carries a backdoor attack by (i) adding a trojan to pattern to his samples, (ii) and mislabeling all his dataset as a chosen target class (see Figure~\ref{fig:trojaned_samples}). We measure the success of adversary's attack (denoted as \emph{backdoor accuracy}) by measuring the accuracy of the model on the poisoned test dataset (whose samples have the trojan pattern, and labeled as the target class chosen by the adversary). \emph{Lower values for backdoor accuracy indicates the model is more robust.}

We present the results for robustness in Table~\ref{tab:robustness}. As can be seen, the trained models become more susceptible to attack as local data distributions become less similar. The results seem more pronounced on Fashion MNIST than on CIFAR10. We believe this is due to the fact that, the attack is already quite successful on $\alpha=1$ CIFAF10, so, there is less room to improve the attack in that setting.

Regardless, we believe the main takeaway of our result is that, any defense mechanism deployed in the FL setting should be tested against different data distributions, in addition to different adversarial agents ratios as is commonly done.

\begin{table}[t]
\large
\centering
\caption{Effect of data distribution on robustness for Fashion MNIST (top), and CIFAR10 (bottom) datasets.}
\begin{tabular}{ c c c } 
 \hline
 Dirichlet $\alpha$ & Accuracy & Backdoor Accuracy \\
 \hline
 $1.0$ & $89.87 \pm 0.14$ & $64.34 \pm 5.12$ \\
  $0.5$ & $89.58 \pm 0.2$ & $69.78 \pm 2.58$ \\
  $0.25$ & $88.47 \pm 0.77$ & $77.34 \pm 2.84$ \\
\end{tabular}
\begin{tabular}{ c c c } 
 \hline
 Dirichlet $\alpha$ & Accuracy & Backdoor Accuracy \\
 \hline
 $1.0$ & $75.72 \pm 0.34$ & $89.66 \pm 0.81$ \\
 $0.5$ & $74.64 \pm 0.13$ & $90.37 \pm 0.4$ \\
 $0.25$ & $71.75 \pm 1.45$ & $91.49 \pm 0.68$ \\
\end{tabular}
\label{tab:robustness}
\end{table}

\subsection{Robustness under a Defense}
Based on our insight from the robustness experiments, we now look at how the robustness is impacted when a defense is deployed under different distributions. To do so, we deploy a recent defense introduced in~\cite{ozdayi2021defending} called \emph{robust learning rate} (RLR). This approach merely modifies the learning rate of aggregation server, per round and per dimension, based on the sign information of agents' updates, and is agnostic to the aggregation function itself.

Concretely, the defense introduces a hyperparameter called the \emph{learning threshold}, denoted as $\theta$ at the server-side.
 For every dimension where the sum of signs of updates is less than $\theta$, the learning rate is multiplied by -1. This is  \emph{to maximize the loss on that dimension rather than minimizing it}. That is, with a learning threshold of $\theta$, the learning rate for the \emph{i}th dimension is given by,
 
\begin{equation*}
\eta_{\theta, i} =  
\begin{cases}
\eta & \left | \sum_{k \in S_t} \sign(\Delta_{t,i}^k) \right | \geq \theta, \\
-\eta  & \text{otherwise.}
\end{cases}
\label{eqn:rlrTheta}
\end{equation*}

For example, consider FedAvg and let $\eta_\theta$ denote the learning rate vector over all dimensions, \ie, $[\eta_{\theta,1}, \eta_{\theta,2}, \dots, \eta_{\theta,d}]^\top$. Then, the update rule with the RLR defense takes the form,
\begin{equation*}
w_{t+1} = w_t + 
\eta_\theta \odot
\frac{\sum_{k \in S_t} n_k \cdot \Delta_t^k}{\sum_{k \in S_t} n_k},
\label{eqn:fedavg_robust}
\end{equation*}
where $\odot$ is the element-wise product operation.

 The experimental results under this defense is presented in Table~\ref{tab:robustnessRLR}. If we compare Table~\ref{tab:robustness} (robustness with no defense), and Table~\ref{tab:robustnessRLR}, we see that the defense diminishes the attack significantly as indicated by the smaller backdoor accuracies. However, at least for CIFAR10 case, the defense becomes substantially weak for $\alpha=0.25$ as evidenced by the low accuracy, and high backdoor accuracy values. Therefore, we can see that local data distributions can substantially tamper with the robustness of the model, even when a defense is deployed.

\begin{table}[t]
\large
\centering
\caption{Effect of data distribution on robustness under RLR defense for Fashion MNIST (top), and CIFAR10 (bottom) datasets.}
\begin{tabular}{ c c c } 
 \hline
 Dirichlet $\alpha$ & Accuracy & Backdoor Accuracy \\
 \hline
 $1.0$ & $88.99 \pm 0.26$ & $0.48 \pm 0.64$ \\
  $0.5$ & $87.96 \pm 0.79$ & $3.62 \pm 2.88$ \\
  $0.25$ & $84.12 \pm 1.4$ & $0.38 \pm 0.44$ \\
\end{tabular}
\begin{tabular}{ c c c } 
 \hline
 Dirichlet $\alpha$ & Accuracy & Backdoor Accuracy \\
 \hline
 $1.0$ & $74.09 \pm 0.54$ & $20.22 \pm 6.66$ \\
 $0.5$ & $69.38 \pm 1.75$ & $13.26 \pm 3.09$ \\
 $0.25$ & $57.07 \pm 3.01$ & $29.89 \pm 20.99$ \\
\end{tabular}
\label{tab:robustnessRLR}
\end{table}

\subsection{Interplay between Fairness and Robustness}
We also look at how a defense might affect the fairness of the model. For example, we see that the RLR defense substantially reduces the backdoor accuracy, but it comes at the cost of some accuracy. We wonder, whether the defense itself can degrade the fairness of the model, and if it does, how this degradation is impacted by the data distributions.
Our experiments with RLR indicates that, the defense itself might make the model less fair, and this effect is increased as local data distributions differ more. 
In Table~\ref{tab:robustnessFairnessNoDefense}, we present the bias of the model under the model poisoning attack without any defense. In Table~\ref{tab:robustnessFairnessRLR}, we present the bias of the model under the model poisoning attack with RLR defense.
As can be seen, we observe that the bias exhibited by the model increases under the defense even though the attack itself is much less successful.

\begin{table}[t]
\small
\centering
\caption{Interplay between fairness and robustness under backdoor attack without any defense for Fashion MNIST (top), and CIFAR10 (bottom) datasets.}
\begin{tabular}{ c c c c } 
 \hline
 Dirichlet $\alpha$ & Accuracy & Backdoor Accuracy & Bias \\
 \hline
 $1.0$ & $89.87 \pm 0.14$ & $64.34 \pm 5.12$ & $29.1 \pm 4.57$ \\
  $0.5$ & $89.58 \pm 0.2$ & $69.78 \pm 2.58$ & $32.8 \pm 10.11 $ \\
  $0.25$ & $88.47 \pm 0.77$ & $77.34 \pm 2.84$ & $33.88 \pm 4.5$ \\
\end{tabular}
\begin{tabular}{ c c c c } 
 \hline
 Dirichlet $\alpha$ & Accuracy & Backdoor Accuracy  & Bias\\
 \hline
 $1.0$ & $75.72 \pm 0.34$ & $89.66 \pm 0.81$ & $35.16 \pm 3.96$ \\
 $0.5$ & $74.64 \pm 0.13$ & $90.37 \pm 0.4$ & $32.02 \pm 4.75 $ \\
 $0.25$ & $71.75 \pm 1.45$ & $91.49 \pm 0.68$ & $56.5 \pm 20.66$ \\
\end{tabular}
\label{tab:robustnessFairnessNoDefense}
\end{table}

\begin{table}[t]
\small
\centering
\caption{Interplay between fairness and robustness under backdoor attack with RLR defense for Fashion MNIST (top), and CIFAR10 (bottom) datasets.}
\begin{tabular}{ c c c c} 
 \hline
 Dirichlet $\alpha$ & Accuracy & Backdoor Accuracy & Bias\\
 \hline
 $1.0$ & $88.99 \pm 0.26$ & $0.48 \pm 0.64$ & $31.84 \pm 5.83$ \\
  $0.5$ & $87.96 \pm 0.79$ & $3.62 \pm 2.88$ & $32.62 \pm 7.78$ \\
  $0.25$ & $84.12 \pm 1.4$ & $0.38 \pm 0.44$ & $51.56 \pm 10.93$ \\
\end{tabular}
\begin{tabular}{ c c c c} 
 \hline
 Dirichlet $\alpha$ & Accuracy & Backdoor Accuracy & Bias \\
 \hline
 $1.0$ & $74.09 \pm 0.54$ & $20.22 \pm 6.66$ & $45.52 \pm 7.18$ \\
 $0.5$ & $69.38 \pm 1.75$ & $13.26 \pm 3.09$ & $63.1 \pm 8.81$ \\
 $0.25$ & $57.07 \pm 3.01$ & $29.89 \pm 20.99$ & $90.26 \pm 1.36$ \\
\end{tabular}
\label{tab:robustnessFairnessRLR}
\end{table}

\section{Discussion and Conclusion}\label{sec:conc}
In this work, we have explored how the differences in local data distributions of the participating agents in a FL setting affect the fairness and robustness properties of the trained models. Our brief experimental analysis have indicated that, just as for accuracy, as the local distributions among agents differ, the fairness and robustness of the trained models degrades. Further, we have showed that these properties might degrade much faster than the accuracy. 

Regardless, our work is a limited exploration of the topic, and we believe there are some avenues for further work. Most importantly, we wonder whether the existing algorithms that are developed to remedy the accuracy drop in \niid settings~\cite{zhao2018federated,fedNonIID2, li2020federated, ozdayi2020improving, shoham2019overcoming, hsieh2020non}, can remedy the fairness and robustness issues as well. Also, the robustness notions we considered in our work was merely against model poisoning attacks. It might be interesting to see how robustness of the models change against privacy attacks~\cite{melis2019exploiting}.

\section*{Acknowledgment}
The research reported herein was supported in part by 
NSF awards, CNS-1837627, OAC-1828467, IIS-1939728, DMS-1925346, CNS-2029661, OAC-2115094 and ARO award W911NF-17-1-0356



%



\clearpage
\bibliography{main}
\end{document}